%% file: fednca.tex
\documentclass[runningheads]{llncs}
\usepackage[T1]{fontenc}
\usepackage{graphicx,verbatim}
\usepackage{hyperref}
\usepackage{booktabs}
\usepackage{subcaption}
\usepackage{svg}
\usepackage{wasysym} 
\usepackage{newunicodechar}
\newunicodechar{€}{\euro}

\usepackage{listings}
\definecolor{backcolour}{rgb}{0.95,0.95,0.92}
\definecolor{codegreen}{rgb}{0,0.6,0}

\usepackage{mathrsfs}
\newcommand{\loss}{\mathscr{L}}
\newcommand{\key}{{\color[RGB]{200,100,30} key}}

\definecolor{red_bright}{HTML}{faf0f0}
\definecolor{red_dark}{HTML}{e5bab9}
\definecolor{green_bright}{HTML}{e2feee}
\definecolor{green_dark}{HTML}{a6dfbf}
\definecolor{blue_bright}{HTML}{deeefd}
\definecolor{blue_dark}{HTML}{afc4db}

\newcommand{\mtheta}{{\color{red_dark} \theta}}
\newcommand{\momega}{{\color{red_dark} \omega}}
\newcommand{\mTheta}{{\color{blue_dark} \Theta}}
\newcommand{\mOmega}{{\color{blue_dark} \Omega}}

\usepackage{algorithm}
\usepackage{algorithmicx}
\usepackage{algpseudocode}
\usepackage[misc]{ifsym}

\algdef{SE}[SUBALG]{Indent}{EndIndent}{}{\algorithmicend\ }%
\algtext*{Indent}
\algtext*{EndIndent}

\AtBeginDocument{%
}

\begin{document}
\title{Equitable Federated Learning with NCA}
\author{Nick Lemke\textsuperscript{1,}\thanks{These authors contributed equally to this work.}, Mirko Konstantin\textsuperscript{\thefootnote}, Henry John Krumb, John Kalkhof, \mbox{Jonathan Stieber}, Anirban Mukhopadhyay}  
\authorrunning{N. Lemke et al.}
\institute{Technical University of Darmstadt, Darmstadt, Germany \\
\textsuperscript{1}Corresponding author: \email{nick.lemke@tu-darmstadt.de}}

\maketitle              
\begin{abstract}
Federated Learning (FL) is enabling collaborative model training across institutions without sharing sensitive patient data. This approach is particularly valuable in low- and middle-income countries (LMICs), where access to trained medical professionals is limited. However, FL adoption in LMICs faces significant barriers, including limited high-performance computing resources and unreliable internet connectivity. To address these challenges, we introduce FedNCA, a novel FL system tailored for medical image segmentation tasks. FedNCA leverages the lightweight Med-NCA architecture, enabling \textbf{training on low-cost edge devices}, such as widely available smartphones, while \textbf{minimizing communication costs}. Additionally, our encryption-ready \mbox{FedNCA} proves to be \textbf{suitable for compromised network communication}. By overcoming infrastructural and security challenges, FedNCA paves the way for inclusive, efficient, lightweight, and encryption-ready medical imaging solutions, fostering equitable healthcare advancements in resource-constrained regions.
We make our implementation publicly available at: \url{https://github.com/MECLabTUDA/FedNCA}

\end{abstract}
\section{Introduction}
Federated Learning (FL) is rapidly emerging as a transformative approach in medical imaging \cite{guan2024federated}. Unlike traditional machine learning methods that rely on centralized datasets, FL facilitates the collaborative training of models across multiple institutions without the need to share sensitive patient data \cite{guan2024federated}. This makes it particularly appealing for applications in healthcare, where adherence to stringent data privacy guidelines, such as HIPAA and GDPR, is paramount \cite{mosaiyebzadeh2023privacy}. 
Even in low-and-middle-income countries (LMICs) with limited resources and a shortage of healthcare professionals, FL has the potential to provide access to high-quality medical AI \cite{nguyen2022federated}. 

While the benefits are evident, the adoption of FL in LMICs is hindered by several infrastructural and technical challenges \cite{ciecierski2022artificial}.
One of the primary barriers in LMICs is the limited access to high-performance computing resources \cite{selvan2024equity}, making it difficult to train deep neural networks and implement traditional FL frameworks. Additionally, slow and unreliable internet connections exacerbate the problem, as the exchange of model updates between clients and servers becomes time-intensive and inefficient \cite{melas2021intrinisic}. 

In state-of-the-art FL solutions, models are becoming increasingly larger, with architectures growing more complex \cite{Liu_FedFMS_MICCAI2024}. The rising computational demands necessitate powerful hardware, stable network infrastructures, and substantial energy resources — requirements that are often inaccessible in resource-constrained regions \cite{he2021fedcv}. As state-of-the-art FL methods evolve, the risk of exacerbating global disparities in AI accessibility increases, underscoring the urgency for lightweight, adaptable, and efficient FL frameworks tailored to LMICs \cite{liu2022no}.

\begin{figure}
    \centering
    \includegraphics[width=\linewidth]{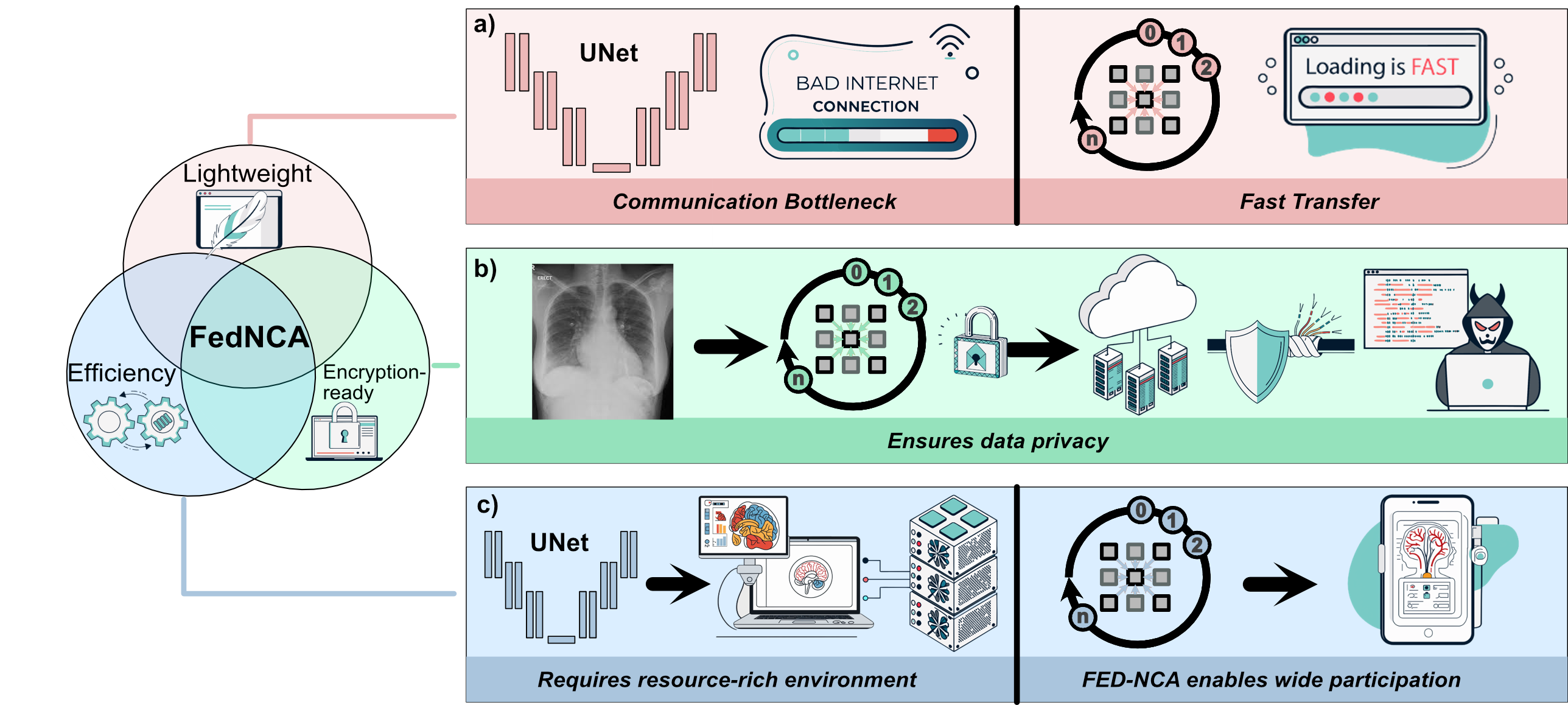}
    \caption{FedNCA a) circumvents problems with low bandwidth internet connections, b) allows for efficient encryption to protect from adversaries or untrusted servers, c) is trainable on diverse hardware including smartphones.}
    \label{fig:motivation_fig}
\end{figure}

To address these challenges, we propose FedNCA, an \textbf{efficient}, \textbf{lightweight}, and \textbf{encryption-ready} FL system for LMICs (Fig.~\ref{fig:motivation_fig}). Unlike conventional FL systems that require substantial computational resources and high-bandwidth connectivity, FedNCA is optimized for deployment in resource-constrained environments.
At its core, FedNCA leverages the Med-NCA backbone~\cite{kalkhof2023med}, which is inherently lightweight and efficient. This approach allows training on low-cost edge devices, such as inexpensive smartphones~\cite{kalkhof2024unsupervised}, which are available even in resource-constrained regions \cite{10.1093/ijpor/edad031}. Moreover, the NCA architecture has $5000\times$ fewer parameters than a UNet, thereby lowering communication costs and enhancing accessibility in regions with limited internet bandwidth. This not only significantly cuts operational costs but also democratizes access to cutting-edge medical AI, allowing broader participation in FL without the barriers of expensive hardware and network constraints.

Another crucial challenge is the presence of untrusted or even malicious servers \cite{jsan12010013}, particularly in rural and remote regions where none of the trusted peers can act as servers. In such scenarios, ensuring data privacy and security becomes a significant concern. Encryption techniques like homomorphic encryption provide strong privacy protection guarantees, even against untrusted or malicious servers \cite{nguyen2023preserving}. However, despite its robust security benefits, homomorphic encryption introduces substantial computational overhead \cite{jin2023fedml}, which can vary depending on the size of the message, potentially limiting its practicality in resource-constrained environments.
Due to the lightweight MedNCA architecture, the \textit{time required for the encryption and decryption process is reduced by a factor of 1800} compared to state-of-the-art architectures like TransUnet \cite{chen2021transunet}. By drastically reducing computational overhead, FedNCA provides strong privacy guarantees, making secure FL more accessible to LMICs. Consequently, \textbf{robust privacy protection is no longer exclusive to those with high computational resources} but becomes available in regions that lack powerful infrastructure.


To address both, the infrastructure and security challenges associated with FL in LMICs we make the following contributions in this work: 1) We propose a secure and communication-efficient FL algorithm specifically for NCAs, 2) We evaluate the trained models in terms of their segmentation performance and the transmission costs coming up during FL, 3) We study the real-world applicability on smartphones, and 4) we analyze the efficiency improvement of the homomorphic encryption on NCAs. 

\section{Background}
\textbf{Neural Cellular Automata}
are characterized by their minimal number of parameters, making them highly efficient compared to traditional neural network architectures. Med-NCA \cite{kalkhof2023med}, an NCA-based architecture specifically designed for medical image segmentation, has demonstrated performance comparable to conventional neural networks, like the UNet \cite{ronneberger2015u} while utilizing only a small fraction of the parameters. 
Med-NCA’s efficiency stems from its iterative application of simple rules. Cellular automata like Conway's Game of Life are known to be Turing-complete, demonstrating that complex computations can be achieved with inherently simple update rules. NCA-based algorithms adapt this idea and encode the rule in a lightweight neural architecture \cite{mordvintsev2020growing}, which significantly reduces hardware requirements compared to large architectures. 
\newline
\textbf{Communication Bottlenecks} in FL arise due to its star-shaped topology, where a central server receives updates from multiple clients and then sends aggregated updates back \cite{wu2024topology}. These updates typically consist of model parameters or gradients, which are often large in size, leading to significant bandwidth consumption and increased latency \cite{melas2021intrinisic}. This challenge is especially pronounced in settings with limited network capacity \cite{yun2024communication}. To mitigate this issue, various encoding and compression techniques have been introduced, such as quantization, sparsification, and federated dropout, which aim to reduce the size of updates while maintaining acceptable model performance \cite{wen2022federated}. However, these methods come with an inherent trade-off: while they decrease communication costs, they may also introduce a performance drop due to the loss of information in the transmitted updates \cite{wilkins2024fedsz}. \newline

\section{Methodology}

\begin{figure}
    \centering
    \includegraphics[width=\linewidth]{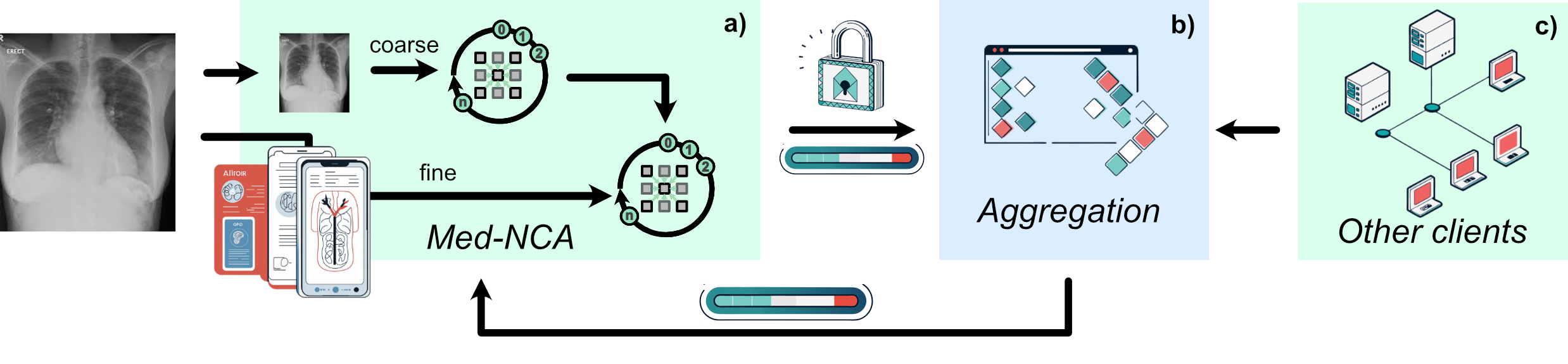}
    \caption{Our FedNCA setup including a) the two-stage Med-NCA backbone, b) the aggregation of encrypted weights, and c) other clients connected via a weak internet connection.}
    \label{fig:method}
\end{figure}

The deployment of secure FL systems in LMICs is often hindered by high communication overhead and computational demands. FedNCA addresses these challenges by providing an \textbf{efficient}, \textbf{lightweight}, and \textbf{encryption-ready} solution.
These key attributes make FedNCA an ideal solution for resource-constrained settings. 
An outline of our FL algorithm can be seen in Fig.~\ref{fig:method} and Alg.~\ref{alg:method}.

\input{code/pseudo}

\textbf{Efficient:} 
Inspired by Med-NCA, each client trains two NCAs via end-to-end backpropagation trough time (BPTT). In detail, the downsampled (coarse) image is given to the first stage, which distributes knowledge on a global scale by running the first NCA $f_\mtheta$ for $T_0=20$ steps. After that, the hidden channels of the NCA are upscaled and concatenated with the high-resolution (fine) image. The second stage uses the global knowledge from the previous NCA to refine the segmentation via a second NCA $f_\momega$ in additional $T_1=40$ steps. During the forward pass, the deep learning engine automatically unrolls the computation of both NCAs $f_\mtheta$ and $f_\momega$ along time (steps) to create a computational graph. 
After computing the cross-entropy loss $\loss$, the gradients are propagated back through time, averaging the gradients of each weight along each time step. Although BPTT is considered to be slow and inefficient, BPTT for NCAs is the opposite. As our NCAs are defined by inherently simple functions $f_\mtheta$ and $f_\momega$, even low-energy devices like smartphones and tablets can train NCAs.

\textbf{Lightweight:} To further enhance communication efficiency, FedNCA’s lightweight architecture ensures that clients transmit only small weight updates to the central server. With its compact 284KB model size, FedNCA drastically reduces bandwidth requirements, making it an ideal fit for FL in low connectivity environments (Fig.~\ref{fig:method} c). This reduces communication costs by nearly $500\times$ compared to traditional U-Net models, mitigating common FL communication bottlenecks. This enables frequent updates even in low-connectivity environments, making it highly suitable for deployment in regions with limited internet access.

\textbf{Encryption-ready:} Homomorphic Encryption (HE) offers a quantum-proof level of security, ensuring privacy preservation even against an untrusted server. This is particularly crucial in FL where reconstruction~\cite{geiping2020inverting} or source inference attacks~\cite{hu2021source} on client updates may leak private information about the client’s training data. HE enables secure aggregation by allowing encryption $\varphi$ and decryption $\varphi^{-1}$ to behave as homomorphisms between plaintext and ciphertext, thereby satisfying 
\begin{equation}
    \varphi(m_1) * \varphi(m_2) = \varphi(m_1 * m_2)\qquad\forall m_1, m_2 \in M
    \label{eq:homomorphism}
\end{equation}
where $M$ represents all possible messages and $*\in\{\cdot, +\}$ represents a group operation, such as addition or multiplication.

As a result, the server can aggregate encrypted client updates without decrypting them, effectively eliminating the threat of server-side data leakage attacks. However, a major downside of HE is its high computational cost, which makes it impractical for large-scale client updates. To mitigate this, FedNCA's low number of parameters facilitates seamless homomorphic encryption, making FedNCA encryption-ready. In our setup, we utilized the CKKS~\cite{cheon2017homomorphic} scheme, which is specifically designed for floating-point numbers, making it highly suitable for FL applications.

By combining efficiency, lightweight design, and encryption-readiness, \mbox{FedNCA} provides a scalable and secure FL solution for LMICs, addressing key barriers to AI adoption in resource-limited settings.

\section{Experiments}

\textbf{Ultrasound:} The \emph{Fetal Abdominal Structures Segmentation}~\cite{corregio2023fetal} dataset includes nearly 1500 images of fetal abdomen circumference (AC). Ultrasound images were captured following a standardized protocol, using Siemens Acuson, Voluson 730 (GE Healthcare Ultrasound), or Philips-EPIQ Elite (Philips Healthcare Ultrasound) systems. For our experiments, we focused on segmenting liver structures in the images. 
For our experiments, we reserve a random set of 118 ($70\%$) patients for the test set. The remaining $51$ patients are randomly split among $5$ FL clients.

\textbf{XRay: } The \emph{MIMIC-III}~\cite{johnson2016mimic} dataset consists of chest XRay images of patients in tertiary care. The segmentation encompasses both the left and right lung. In our experiments we utilize a random subset of 50 images, reserving 25 images for the test set and distributing the other 25 evenly among 5 clients.

\textbf{Baselines: } We compare FedNCA to a UNet~\cite{ronneberger2015u} and TransUNet~\cite{chen2021transunet}. Additionally, we use quantization and sparsification methods, to reduce the transmission cost in the federated protocol. Specifically, we quantize the model weights to floating point values with $4$-bit precision, indicated by \emph{4-bit}.
Furthermore, we sparsify the model weights sent to the server by an unidirectional top-k algorithm~\cite{han2020adaptive}. Our algorithm selects only the most important client parameters, discarding the others. We consider parameters to be important by their difference to the values they had in the last FL round. We select the largest $k\%$ of parameters to be sent to the server in upstream. The server, on the other hand, sends all aggregated parameters to the clients. We indicate this method by \emph{top-k}, where $k\%$ indicates the number of parameters sent in each upstream. 

\textbf{Quantitative Scores:} We measure the segmentation precision of each method using the \emph{Dice} score. The \emph{transmission cost} measures the amount of data (in MiB) transferred in each FL round including the model parameters and the associated metadata introduced by the quantization or sparsification algorithms.


\section{Results}
In this section, we present our results when training in low-bandwidth regions. We investigate runtime when training on affordable hardware, and we measure runtime for encrypting weights of different segmentation models.

\begin{figure}
    \centering
    \includegraphics[width=\linewidth]{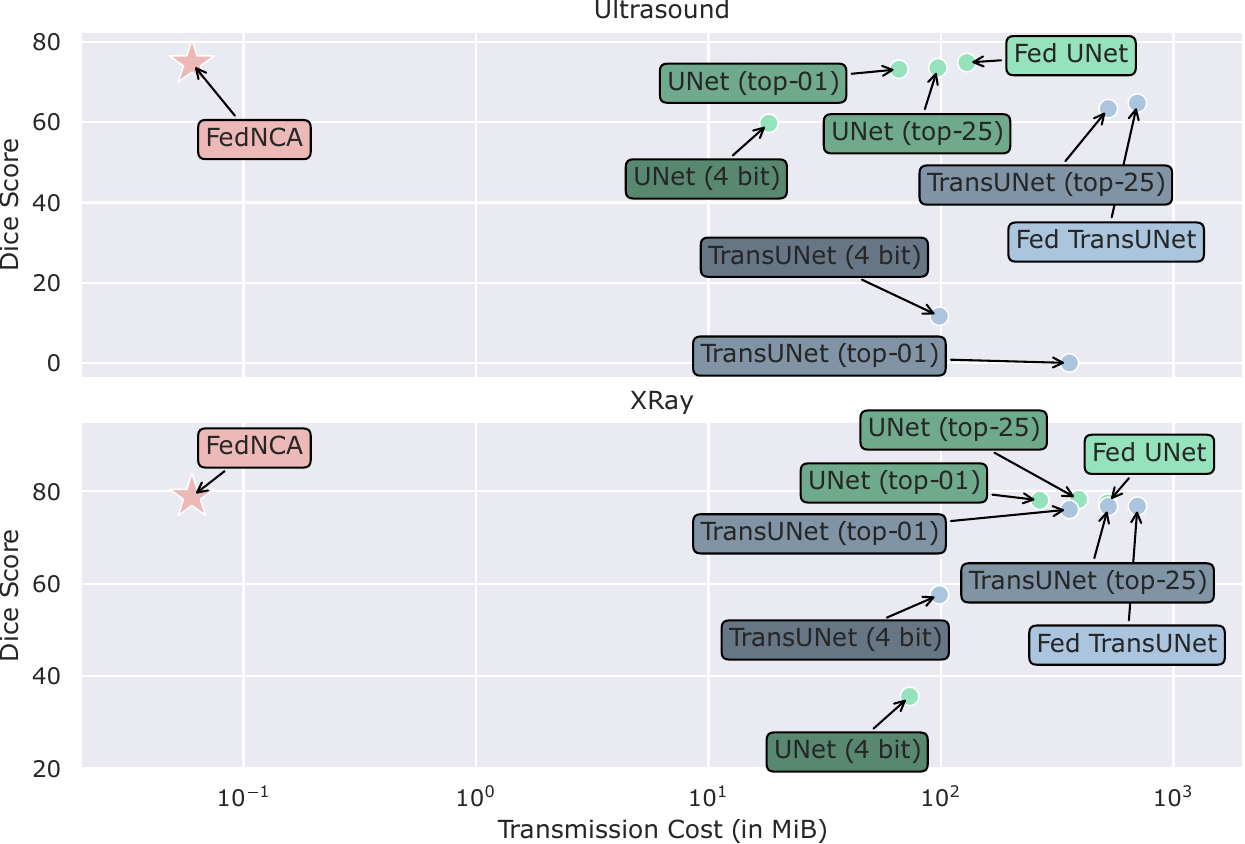}
    \caption{Dice score and transmission cost in MiB. FedNCA achieves the best Dice while requiring the least transmission cost.}
    \label{fig:dice_cost}
\end{figure}

\textbf{Training in low-bandwidth regions:}
The quantitative results presented in Fig.~\ref{fig:dice_cost} provide insights into the Dice score and transmission costs of the model parameters. The results demonstrate that FedNCA consistently achieves equal segmentation quality of $74\%$ and $78\%$ to the baseline models, despite its \emph{$2000\times$ lower communication overhead}. To address these high transmission costs, we apply compression techniques to the U-Net-based approaches. However, \emph{even after compression, the transmission costs remain at least $300\times$ higher} than those of FedNCA, indicating that our method is inherently more efficient without requiring additional compression. Furthermore, the compressed model updates experience significant performance degradation, revealing a clear tradeoff between reducing communication costs and maintaining model accuracy. 
FedNCA, on the other hand, is much more efficient without any decline in segmentation accuracy.

\begin{figure}
    \centering
    \includegraphics[width=\linewidth]{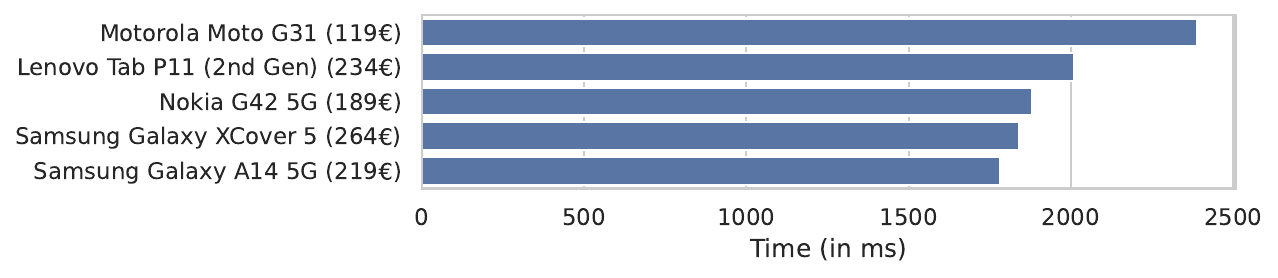}
    \caption{Training time per epoch of FedNCA on heterogeneous hardware.}
    \label{fig:training_time}
\end{figure}

\textbf{Training time on inexpensive hardware:}
We perform a case study to investigate the trainability of FedNCA on affordable devices. 
We report the time taken for training a single epoch on heterogeneous hardware. The results in Fig.~\ref{fig:training_time} show that training on smartphones and tablets cheaper than 300\euro{}  is feasible. 

\begin{figure}
    \centering
    \includegraphics[width=\linewidth]{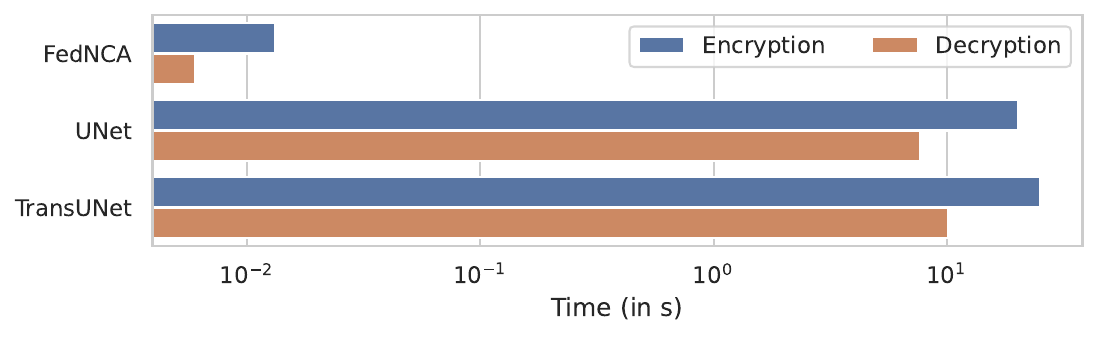}
    \caption{Time (in s) taken for homomorphic encryption and decryption of FedNCA, UNet, and TransUNet on an Intel i7-13700K CPU.}
    \label{fig:encrypt_time}
\end{figure}

\textbf{Efficiency of homomorphic encryption:}
To preserve privacy during federated training, even in the case of an untrusted server, we encrypt the parameters using a homomorphic encryption scheme. In Fig.~\ref{fig:encrypt_time} we report a runtime analysis of the homomorphic encryption and decryption on FedNCA, UNet, and TransUNet on a recent Intel CPU. While, the encryption scheme adds less than $20$ milliseconds for FedNCA, encryption and decryption of the UNet is \emph{$1400\times$ slower}, adding $27$ seconds to each FL round. For the bigger \mbox{TransUNet}, the runtime is \emph{$1800\times$ longer}, rendering FL slow and inefficient, especially for low-cost computing machinery.


\section{Conclusion}

In this work, we introduced FedNCA, a FL framework that leverages Neural Cellular Automata-based architecture for \textbf{efficient}, \textbf{lightweight}, and \textbf{encryption-ready} model training. 
Our experiments demonstrated robust performance across two segmentation tasks while showcasing efficient communication, enabled by the lightweight model architecture. Additionally, the low parameter count allowed the models to be trained on inexpensive smartphones, significantly reducing the participation burden within this framework. Furthermore, the reduced number of parameters enhances the efficiency of Homomorphic Encryption, making our solution encryption-ready and well-suited for secure, privacy-preserving applications.
Traditional model architectures, such as U-Net and TransUNet, utilize compression methods to alleviate the communication bottleneck in FL, making it more accessible in regions with limited internet bandwidth. However, these compression techniques come with a tradeoff in performance. In contrast, FedNCA has been shown to be more communication-efficient while outperforming compressed versions of these models.
The challenge of using encryption in traditional solutions arises from the high computational cost of encryption and decryption. Our results demonstrate that, due to its model architecture, FedNCA significantly reduces this computational burden, making it encryption-ready for real-world applications.
FedNCA allows \emph{equitable participation in AI training}, enabling even low-resource clinics and underrepresented populations to contribute without the requirement of data sharing. The tiny architecture runs on widely available devices, including smartphones, making FL accessible where traditional models are impractical. By combining efficient segmentation with federated training, Med-NCA ensures AI models benefit from diverse global data, making high-quality medical AI available to all, regardless of infrastructure or location.

\begin{credits}
\subsubsection{\ackname}
This work has been partially funded by the Federal Ministry of Education and Research project ”FED-PATH” (grant 01KD2210B) and by the \mbox{hessian.AI} project "Fed-NCA".

\subsubsection{\discintname}
The authors have no competing interests to declare that are relevant to the content of this article.
\end{credits}

\bibliographystyle{unsrt}
\bibliography{fednca}
\end{document}

%% file: code/pseudo.tex
\begin{algorithm}
\caption{
FedNCA client and server functions.
}
\label{alg:method}
\begin{algorithmic}
\State 
\begin{itemize}
    \item $\varphi$: Encryption function with corresponding \key
    \item $f_\mtheta$, $f_\momega$: Backbone NCAs with weights $\mtheta, \momega$  and encrypted weights $\mTheta,\mOmega$
    \item $\loss$: Segmentation loss function, $\eta$: Learning rate
    \item $x$, $y$: Image $x$ and segmentation target $y$ (for simplicity here only one)
    \item $T_0, T_1$: Number of steps for fine and course NCA
\end{itemize}


\State \textbf{ClientUpdate}($x$, $y$, $\{\mOmega, \mTheta\}$):
\Indent
\State $\{\mtheta, \momega\}\gets\varphi^{-1}(\{\mTheta,\mOmega\}$, \key$)$\Comment{Decrypt parameters}
\State $z\gets$ \textbf{downscale}($x$)
\For{each step $s = 0...T_0$}
    \State $z\gets f_\mtheta(z)$
\EndFor
\State $z\gets$\textbf{upscale}$(z,x)$
\For{each step $s = 0...T_1$}
    \State $z\gets f_\momega(z)$
\EndFor
\State $\mtheta\gets\mtheta-\eta\nabla_\mtheta\loss(x,y)$\Comment{Compute loss and update parameters}
\State $\momega\gets\momega-\eta\nabla_\momega\loss(x,y)$
\State Return $\varphi(\{\mtheta, \momega\},$ \key$)$ to the server\Comment{Encrypt and return parameters}
\EndIndent


\State \textbf{ServerUpdate}($\{\mOmega_0, \mTheta_0\}$, ..., $\{\mOmega_n, \mTheta_n\}$):\Comment{Aggregate parameters from $n$ clients}
\Indent
\State $\mTheta\gets\frac{1}{n}(\mTheta_0$ + ... + $\mTheta_n$)\Comment{Average encrypted parameters}
\State $\mOmega\gets\frac{1}{n}(\mOmega_0$ + ... + $\mOmega_n$)
\State Return $\{\mTheta,\mOmega\}$ to the clients
\EndIndent

\end{algorithmic}
\end{algorithm}